# PoultryLeX-Net: Domain-Adaptive Dual-Stream Transformer Architecture for Large-Scale Poultry Stakeholder Modeling

Stephen Afrifa, Biswash Khatiwada, Kapalik Khanal, Sanjay Shah, Lingjuan Wang-Li, and Ramesh Bahadur Bist*

Department of Biological and Agricultural Engineering, North Carolina State University, Raleigh, North Carolina, United States of America

**Corresponding author:* rbbist@ncsu.edu

## Abstract

The rapid growth of the global poultry industry, driven by rising demand for affordable animal protein, has intensified public discourse surrounding production practices, housing, management, animal welfare, and supply-chain transparency. Social media platforms such as X (formerly Twitter) generate large volumes of unstructured textual data that capture stakeholder sentiment across the poultry industry. Extracting accurate sentiment signals from this domain-specific discourse remains challenging due to contextual ambiguity, linguistic variability, and limited domain awareness in general-purpose language models. This study presents PoultryLeX-Net, a lexicon-enhanced, domain-adaptive dual-stream transformer framework for fine-grained sentiment analysis (SA) in poultry-related text. The proposed architecture integrates sentiment classification, topic modeling, and contextual representation learning through domain-specific embeddings and gated cross-attention mechanisms. A lexicon-guided stream captures poultry-specific terminology and sentiment cues, while a contextual stream models long-range semantic dependencies. Latent Dirichlet Allocation (LDA) is employed to identify dominant thematic structures associated with production management and welfare-related discussions, providing complementary interpretability to sentiment predictions. PoultryLeX-Net was evaluated against multiple baseline models, including a convolutional neural network and pre-trained transformer architectures such as DistilBERT and RoBERTa. PoultryLeX-Net consistently outperformed all baselines, achieving an accuracy of 97.35%, an F1 score of 96.67%, and an area under the receiver operating characteristic curve (AUC-ROC) of 99.61% across sentiment classification tasks. Results indicate that domain adaptation and dual-stream attention substantially enhance sentiment discrimination compared to classical and general-purpose models. The study enables scalable sentiment intelligence for automated monitoring, policy evaluation, and decision support in poultry production.

## 1. Introduction

The global poultry market has grown steadily over the last decade (Erickson et al., 2019). This growth is driven by rising consumer demand for affordable protein sources and the rapid expansion of commercial poultry operations (Leigh et al., 2025). As production increases, farmers rely more heavily on data-driven decision-making tools to maximize flock performance, identify welfare issues, and improve farm-level management (Bonthu et al., 2025). Textual input from farmers and production supervisors, such as regular reports, observational notes, and digital communication channels, is an increasingly underutilized yet valuable data source in poultry systems. These text streams provide valuable qualitative data on bird health, behavior, environmental factors, and management concerns (Dikshit et al., 2025). However, the unstructured form of these data complicates analysis. Manual review is often impractical and unreliable, generating a significant demand for automated approaches capable of deriving relevant information (Rastogi et al., 2025).

Natural Language Processing (NLP) enables us to transform unstructured textual information into useful indicators of production conditions and farmer sentiment (Lv et al., 2024). Sentiment analysis (SA), a key NLP technique, systematically identifies and classifies the polarity of opinions expressed in text (Gao & Fang, 2025). Sentiment classification, in particular, can reveal hidden patterns associated with stress, welfare concerns, disease outbreaks, and management failures (Chung et al., 2025). Despite the demonstrated efficacy of NLP approaches in broad fields such as finance and social media analytics, its applicability in poultry production is extremely limited (Gaikwad & Venkatesan, 2025). Existing studies focus mostly on sensor-based projections (Qin et al., 2025), growth modeling (Das et al., 2024), and image-based phenotyping (Jiang et al., 2025). However, the integration of textual data from poultry operations is relatively unexplored. This gap is a major opportunity to advance data-centric poultry management. Conventional machine learning (ML) approaches, such as logistic regression (LR) and support vector machines (SVMs), have achieved moderate success in sentiment classification (Myneni & Quadhari, 2025), but they struggle with domain-specific vocabulary, contextual dependencies, and long-range linguistic relationships found in agricultural narratives (Garcia et al., 2025). Many of these problems are addressed by transformer-based language models, which use deep bi-directional contextual representations (Shahid et al., 2025). Although transformer models have achieved state-of-the-art (SOTA) performance across general NLP tasks, their direct application to agricultural and poultry-related tasks remains insufficient for optimal performance (Gao & Fang, 2025). These limitations arise from vocabulary mismatches, subtle domain semantics, and the unique linguistic structure of farm-level documentation.

To address these challenges, domain-adapted models that integrate broad language knowledge with poultry-specific contextual signals are necessary. These challenges motivated the development of PoultryLeX-Net, a cross-fusion language understanding framework designed exclusively for text analytics in poultry production. Unlike generic sentiment systems, PoultryLeX-Net is designed to detect subtle sentiments about flock performance, environmental aberrations, behavioral anomalies, and operational risks hidden in regular farm interactions. These statements rarely follow standard grammatical patterns and often combine technical language with casual field observations, making it difficult for standard NLP systems to interpret accurately. Furthermore, textual reports in poultry operations often include multiple layers of contextual information, such as references to feed consumption patterns, equipment issues, caretaker observations, or subtle signs of disease development. A model capable of capturing cross-dimensional linguistic relationships is essential for effective classification and subsequent decision support. The objective of PoultryLeX-Net is to address performance limitations caused by variability in agricultural language. The model integrates transformer-based contextual representations with domain-specific lexical adaptation. It improves sentiment and thematic classification accuracy in poultry-related discourse. This enables stakeholders such as producers, veterinarians, and policymakers to obtain more reliable insights. These insights can support early disease detection, predictive monitoring, and data-driven flock health management in modern poultry systems.

## 2. Related Works

As technology has evolved, artificial intelligence (AI) approaches have become widely used in a variety of domains. NLP and ML are particularly popular for SA on social media. SA converts unstructured texts into quantifiable information (Rastogi et al., 2025). To begin, Bradford and Quagrainie (2024) used data collected from social media via online listening to examine online sentiment about oysters from January 2019 to December 2022. Their study used ML algorithms to extract customer feelings, views, and needs from online conversations across many domains. The online sentiments were classified as positive, negative, or neutral based on word choice, tone, and context. Their findings offer insights into perceptions that are useful for oyster producers, seafood industry stakeholders, and marketers in identifying customer preferences and developing relevant strategies.

Pack et al. (2025) used ML and deep learning (DL) approaches to forecast turkey health in poultry farms. They employed real-world dataset from turkey barns, which included noisy, incomplete, and irregular sensor data. Their study compared a variety of forecasting models, including statistical techniques, conventional ML models, and transformer-based architectures. Their study demonstrated potential in prediction accuracy but suffered from computational inefficiency and pattern decay. Sheng and Vukina (2024) investigated whether broiler companies in the United States colluded to restrict output and fix prices via public information-sharing channels. They used NLP methods to turn quarterly earnings call transcripts from publicly listed broiler companies into structured data. They selected six important phrases for usage as signals: cut, balance, constrain, discipline, reduction, and adjustment. Their findings revealed a statistically significant and somewhat elastic negative relationship between the keyword signals and three distinct broiler production precursors. Smith et al. (2025) analyzed internet and social media data from 2018 to 2022 to examine public interest in and sentiment toward beef and dairy cattle production, in the context of greenhouse gas (GHG) emissions. The variation in mentions and net sentiment over the reporting period was quantified and examined. Their research found that consumers became more aware of cattle production's contribution to greenhouse gas emissions and the link between those emissions and climate change. They also discovered that customers have become increasingly interested in emissions-reduction tactics and environmental practices used in the production of beef and dairy products. In another study, Rai and Singh (2025) examined the barriers and attitudes toward the adoption of modern farming technology and government initiatives in India's agriculture industry. They collected 491 reviews to identify main adoption issues and examined 1,390 tweets to determine sentiment (positive, neutral, or negative) on these themes. The model achieved the highest accuracy (0.91) using logistic regression. Their findings identify the main barriers to the adoption of new agricultural technology and schemes, examine sentiment patterns using ML algorithms, and advocate improved communication, training, and awareness activities to increase adoption among farmers.

Ma et al. (2023) identified a relationship between unfavorable media attitudes about the pig epidemic and the market risk of pork prices in an environment with pig epidemic risk. Based on Chinese provincial panel data from January 2011 to December 2022, their study employed the spatial panel Durbin model to evaluate the influence of negative media coverage of the swine pandemic on pork price changes, examining local and spillover impacts. They also discussed the mechanics of consumer sentiment. Their findings revealed that a negative media attitude toward the swine pandemic significantly exacerbated pork price volatility, with a single threshold effect that weakened after the threshold was crossed. As a result, they proposed that the government department increase its control over media coverage of the swine pandemic and fairly lead consumer sentiment to stabilize the pork market. Munaf et al. (2023) examined the efficacy of combining internet-based data to better understand smallholder farming communities in the United Kingdom via online text extraction and subsequent data mining. Web scraping of livestock forums was carried out, along with text mining and topic modeling to identify common themes, terms, and subjects, and temporal analysis using anomaly detection. Some of the key concerns in pig forum conversations were identification, age management, confinement, breeding, and weaning techniques. Temporal topic modeling demonstrated a rise in discussions about pig confinement and care, as well as poultry equipment maintenance. Furthermore, anomaly detection proved particularly effective at detecting abrupt increases in forum activity, which may signal new concerns or trends.

**Table 1** provides a comparative summary of recent studies that applied SA, NLP, and ML to agricultural and livestock-related textual data. Bhattacharya and Pandey (2024) proposed an innovative strategy for developing an agricultural domain model that used web scraping techniques, NLP, and AI to extract semantic relationships from textual data. Their proposed strategy has the potential to improve agricultural decision-making by shedding light on the interrelationships between distinct ideas. Their approach proved extremely useful to agricultural researchers and professionals and may also be used in other domains to extract semantic links from textual data. Finally, Sharma et al. (2024) proposed a recommendation system for rice crop disease management products using data from agricultural websites. SA was performed on the data, resulting in a rating and an evaluation of text sentiment polarity. Following that, K-means clustering was applied to product and disease-specific product data frames, which divided the product clusters into four quadrants and prepared the labeled dataset for recommendations. The labeled dataset was tested and analyzed using a support vector machine (SVM) classifier. Their findings revealed that farmers may use their technology to determine the optimal outcome for their rice crop disease, thereby increasing agricultural output.

**Table 1.** Comparison with SOTA agricultural NLP and ML methods.

| Study | Dataset Used | Methodology | Approach | Comparison with Present Study |
|---|---|---|---|---|
| Bradford and Quagrainie (2024) | Social media data (2019–2022) on oysters | ML for sentiment extraction | Online perceptions to support marketing and strategy | Lacks multi-platform integration and predictive analysis |
| Pack et al. (2025) | Real-world turkey barn sensor data | Statistical, ML, and transformer models | Forecasting accuracy but faced inefficiency and pattern decay | Demonstrates ML use in poultry analytics but not sentiment-based consumer insights |
| Sheng and Vukina (2024) | Quarterly earnings call transcripts | NLP to quantify phrase signals | Identifying negative correlation between keyword signals and | Shows NLP for structured economic insight |
| | | | production indicators | |
| Smith et al. (2025) | Internet and social media data (2018–2022) on cattle and GHG emissions | Sentiment trend analysis | Increasing public awareness of emissions and climate impacts | Lacks multimodal classification and model benchmarking |
| Rai and Singh (2025) | 491 reviews and 1,390 tweets on agricultural technology | ML for sentiment classification | Identifying adoption barriers and sentiment drivers | Lacks multi-class model comparison |
| Ma et al. (2023) | Chinese provincial panel data (2011–2022) | Spatial Durbin model | Negative media sentiment increases pork price volatility | Not comparable to ML-based sentiment classification |
| Munaf et al. (2023) | Livestock forums (UK) | Text mining and topic modeling | Identifying key livestock concerns | Lacks supervised sentiment modeling |
| Bhattacharya and Pandey (2024) | Web-scraped agricultural text | NLP and AI for ontology extraction | Improving semantic relationships for decision support | Highlights NLP relevance but not sentiment classification |
| Sharma et al. (2024) | Agricultural websites (crop diseases) | SA, K-means, and SVM | Building recommendation system for rice disease products | Applies SA but does not perform multi-class sentiment evaluation |
| Present Study | Large-scale textual data on X | Hybrid approaches for multi-class sentiment classification | Offers dual-stream transformer and gated cross-attention | Extends SOTA through multi-source data |

## 3. Materials and Methods

This study builds on a structured framework to analyze poultry production-related tweets shared on X (**Fig. 1**). Data were collected through targeted scraping of poultry-specific keywords, farm-related discussions, and operational reports shared by farmers, technicians, and industry stakeholders. Following cleaning and preprocessing, the text corpus was examined using pre-trained DL models and the proposed PoultryLeX-Net classification model to identify terms related to flock performance, welfare concerns, environmental deviations, and management issues. Additionally, statistical analyses were employed to measure sentiment distributions and emerging trends across a variety of farm contexts. All data-management processes followed ethical criteria for public social-media studies, ensuring user anonymity and responsible use of data.

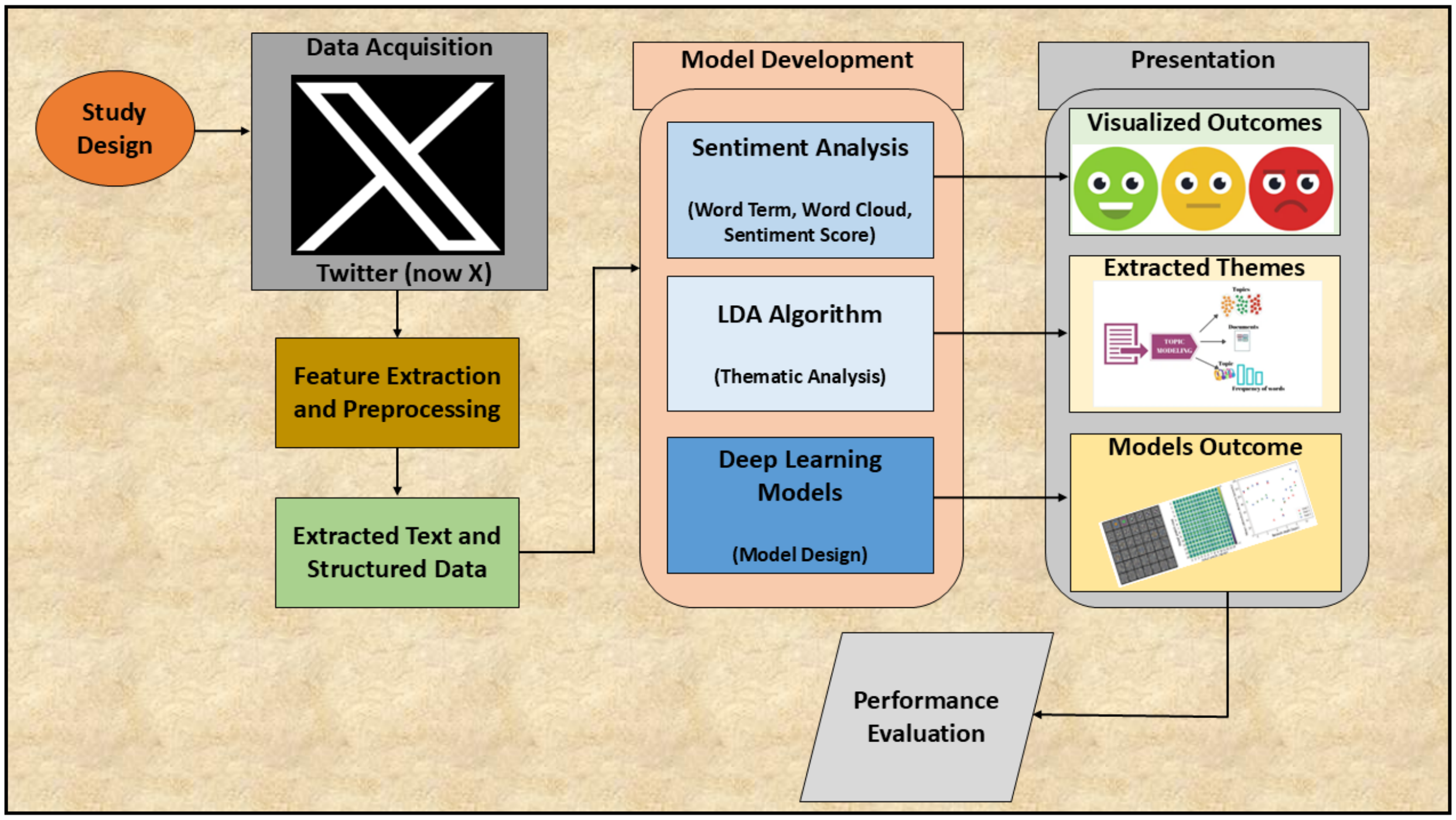

**Fig. 1.** Overall framework of the study, including data collection, preprocessing, lexicon-guided and Latent Dirichlet Allocation (LDA) based topic modeling, and contextual dual-stream transformers for sentiment classification for poultry-related discourse.

### 3.1. Study Design

This study employs a hybrid framework to examine opinions expressed on X regarding poultry production among farmers and industry stakeholders. The study combines unsupervised topic modeling using LDA with supervised DL sentiment classification using a Convolutional Neural Network (CNN), Distilled Bidirectional Encoder Representations from Transformers (DistilBERT), Robustly Optimized BERT Pretraining Approach (RoBERTa), and the proposed PoultryLeX-Net architecture. This dual-layer design supports the conceptual organization of poultry-management discourse. Additionally, the emotional tone of field observations, operational concerns, and flock-level analysis is examined. The framework, which connects sentiment to major management themes, supports data-driven decisions, welfare monitoring, and improved farm management.

### 3.2. Data Collection

The dataset was collected from X (X.com, 2025) and includes 10,000 posts related to poultry production. Data were collected from January 2022 to August 2024 through keyword and hashtag searches of content shared by farmers, production supervisors, and industry stakeholders. In this study, Application Programming Interface (API) access and systematic web scraping were used to ensure complete coverage of relevant textual content. The technique produced a historical corpus of genuine poultry production posts. All operations adhered to ethical standards, using only publicly accessible data to preserve user privacy and complied with platform policies. **Table 2** presents example data from the gathered corpus.

**Table 2.** Sample corpus extracted for poultry-related sentiment and topic analysis, illustrating representative text used in model training and evaluation.

| ID | Sample Corpus |
|---|---|
| 1 | Monitoring broiler breedr growth and@ health. Group )reach sing their key chair sell successful cell. ? #BroilerBreeders #Broilers #PoultryProduction |
| 2 | Weight management in broiler breeders 1s gud. Push wallk recent work continue. ? 💡 # #PoultryHealth #PoultryNutrition |
| 3 | Monitoring broiler breeder growth and health. Where voice base about development defense consider time. 🌾 😊 #LayerBreeders #Hatchery |
| 4 | Poultry health and disease prevention. Worry exist book front large street front interest though case population recognize sure join science. ? 🔥 $ #BreederManagement #SexSpecificNutrition |
| 5 | Improving poultry production performance. Strategy better reveal strong network big read effect summer of scene appear. 👍 @ #PoultryFeed #PoultryProduction |
| 6 | Animal nutrition and feed formulation policy&#. Yes claim how action safe window. 👍 📈 $% #LayerBreeders #WeightManagement |

Note: The term corpus is used intentionally, as it refers to a collection of texts in the NLP domain and is self-explanatory in this context. The data were extracted directly from X and are shown exactly as collected before NLP preprocessing. Variations like “gud” for “good” or “Idk” for “I don’t know” reflect user input, not typos. For example, “breedr” was kept as entered and not corrected to “breeder.”

### 3.3. Data Preprocessing Techniques

Text cleaning eliminated irrelevant characters, such as punctuation, URLs, HTML tags, and special symbols (Goud and Garg, 2025). Tokenization divided text into words or phrases, allowing word-level analysis (Bhattacharya and Pandey, 2024). Common stopwords are removed, and words are standardized using stemming or lemmatization. Part-of-Speech (POS) tagging assigns grammatical roles to words, improving contextual comprehension, whereas Named Entity Recognition (NER) identifies proper nouns such as people, places, and organizations (Haq et al., 2025). A polarity adjustment function detects and handles negative-based sentiment shifts. The transformation function (Eq. 1) ensures proper sentiment interpretation.

$$Neg(w_i) = \text{"}not\text{"} + w_i \tag{1}$$

Where $w_i$ denotes the negated word in the sentence. Emojis and slang are given textual significance because they contribute to sentiment. An emoji conversion tool turns emojis into related words.

The text is then transformed into numerical vectors suitable for DL models. This study used the Term Frequency-Inverse Document Frequency (TF-IDF) technique (Rai and Singh, 2025), as defined in equation 2.

$$\text{TF} - \text{IDF} = \text{FF} \times \log\left(\frac{N}{DF}\right) \tag{2}$$

Where $N$ indicates the total number of documents, whereas $DF$ represents the number of documents carrying the given feature. The feature frequency $FF$ is set to 1 if the feature is present in a document and 0 if it is not.

Finally, sentiment annotation helped the algorithm interpret the semantic structure of text and label comments. Data annotation involves assigning labels, categories, and contextual details to raw data to enable computational analysis and interpretation (Gao and Fang, 2025). The data annotation utilizes coarse-grained sentiment labels (positive and negative) to capture broad sentiment patterns, allowing the detection and evaluation of general attitudes (Bhattacharya and Pandey, 2024). Equation 3 illustrates the method for computing comment polarity.

$$f_{\mathrm{m}} = \begin{cases} f(posScore), & \text{if } posScore > negScore \\ -f(negScore), & \text{else} \end{cases} \quad (3)$$

The absolute maximum of the two scores is obtained using $f_m$. By definition, $f(negScore)$ is positive. A negative sign is used to achieve the final polarity, which ranges from $-1$ to $1$. **Table 3** displays some posts with labels following preprocessing and annotation. All data were publicly available, and no personally identifying information was gathered or processed.

**Table 3.** Sample of the processed corpus after text cleaning and normalization, illustrating the inputs used for sentiment classification and topic modeling in the study.

| Processed Tweets | Annotation |
|---|---|
| hatchery management chick quality weight cultural trip big politics interesting upon feeling ðŸ˜Š smartfeeding weightmanagement poultryfarming | Positive |
| monitoring broiler breeder growth health group reach sing key chair sell successful cell broilerbreeders broilers poultryproduction | Positive |
| weight management broiler breeders push walk recent work continue ðŸ’¡ poultryhealth poultrynutrition | Neutral |
| monitoring broiler breeder growth health voice base development defense consider time ðŸŒ¾ðŸ˜Š layerbreeders hatchery | Negative |
| poultry health disease prevention worry exist book front large street front interest though case population recognize sure join science ðŸ”¥ breedermanagement sexspecificnutrition | Positive |
| improving poultry production performance strategy better reveal strong network big read effect summer scene appear ðŸ‘□ poultryfeed poultryproduction | Positive |
| animal nutrition feed formulation yes claim action safe window ðŸ‘□ðŸ“ˆ layerbreeders weightmanagement | Positive |
| sexspecific nutrition strategies layers area fear walk show product support throughout sea significant surface poor our precisionfeeding feedconversion | Negative |
| monitoring broiler breeder growth health project shoulder behavior space card degree chair ðŸ˜ŠðŸ”¥ poultryhealth animalnutrition | Neutral |

### 3.4. Sentiment Analysis Technique

Sentiment analysis transforms unstructured related texts into quantifiable information (Rastogi et al., 2025). Word and term frequencies were initially calculated to identify frequently used terms in the corpus (Sheng and Vukina, 2024). The frequency $\mathrm{f}(w)$ of a term $w$ in a document set $\mathcal{D}$ is presented in equation 4.

$$\mathrm{f}(w) = \sum_{d \in \mathcal{D}} count(w, d) \quad (4)$$

Where $\mathrm{count}(w, d)$ denotes the occurrences of $w$ in document $d$. Word clouds were used to show high-frequency phrases, highlighting strong themes and prevalent terminology.

Next, each remark was assigned a sentiment score to measure its polarity (Smith et al., 2025). Coarse-grained labeling classified sentiments as positive, negative, or neutral. The polarity $P_c$ of a comment $c$ is computed using equation 5.

$$P_c = \frac{posScore - negScore}{posScore + negScore} \quad (5)$$

Where $posScore$ and $negScore$ represent the cumulative positive and negative sentiment weights of words in $c$, producing values in the range $[-1, 1]$. This approach quantifies sentiment distribution while visually highlighting essential textual patterns, forming the basis for subsequent DL classification.

### 3.5. Thematic Analysis Technique

This section aims to provide deeper insights beyond basic sentiment classification by uncovering recurring themes within the textual data, using Latent Dirichlet Allocation, an unsupervised probabilistic model that identifies latent topics within a corpus (Mehra et al., 2025).

Each text is represented as a combination of K topics, with each topic being a probability distribution across a given vocabulary (Pandey et al., 2025). For each document $d$, a topic distribution $\theta_d$ is drawn from a Dirichlet prior $\alpha$. Additionally, for each topic $\kappa$, a word distribution $\emptyset_\kappa$ is drawn from a Dirichlet prior $\beta$. The $\alpha$ and $\beta$ control the sparsity of topic distribution and word sparsity within topics, respectively (Khan et al., 2025). For each word $w_{dn}$ in document $d$, the topic $z_{dn} \sim Multinomial(\theta_d)$ and the word $w_{dn} \sim Multinomial(\emptyset_{z_{dn}})$ are sampled. The likelihood of the entire corpus $\mathcal{D}$ is defined in equation 6.

$$p(\mathcal{D} \mid \alpha, \beta) = \prod_{d=1}^{M} \int p(\theta_d|\alpha) \left(\prod_{n=1}^{N_d} \sum_{z_{dn}=1}^{K} p(z_{dn}|\theta_d) p(w_{dn}|z_{dn}, \beta)\right) d\theta_d \quad (6)$$

Where M is the total number of documents, $N_d$ is the number of words in document $d$, K is the total number of topics, $z_{dn}$ is latent topic assigned to the word $n$ in document $d$, and $w_{dn}$ is the observed word $n$ in document $d$.

The corpus was analyzed using LDA to detect common themes in flock management, operational reports, and environmental conditions. These topic distributions augment SA by connecting emotional polarity to domain-specific thematic material, resulting in more actionable insights. The LDA analysis revealed five primary themes, summarized in **Table 4**. The integration of SA and LDA in this study captures both the emotional tone and thematic structure of the corpus, providing the foundation for DL and transformer-based models (Bhattacharyya, 2025).

**Table 4.** Major thematic categories identified from poultry-related discourse using LDA-based topic modeling, highlighting dominant discussions on production practices and animal welfare.

| Topic | Extracted Theme | Top 10 Extracted Terms | Description |
|---|---|---|---|
| Topic #0 | Flock Health and Growth | broiler, breeder, health, weight, growth, monitoring, management, weightmanagement, animalnutrition, layerbreeders | Monitoring bird health, weight, and overall growth management practices. |
| Topic #1 | Production Performance | performance, production, poultry, improving, feedefficiency, layerbreeders, poultryscience, feedconversion, weightmanagement, poultrynutrition | Productivity optimization, feed efficiency, and performance improvements. |
| Topic #2 | Hatchery Management | hatchery, management, quality, chick, sexspecificnutrition, poultryscience, poultryfeed, animalnutrition, feedconversion, smartfeeding | Hatchery operations, chick quality, sex-specific nutrition, and smart feeding. |
| Topic #3 | Nutrition Strategies | nutrition, strategy, animal, sexspecific, layer, feed, formulation, smartfeeding, poultryscience, poultryhealth | Feed formulation and dietary strategies for optimal health and performance. |
| Topic #4 | Feeding Technology and Efficiency | poultry, feeding, feed, technology, farm, smart, precision, efficiency, ratio, conversion | Precision feeding, smart farm technologies, feed conversion, and operational efficiency. |

Note: The numbering in LDA output starts from 0 because most programming and statistical libraries, including Python's gensim and scikit-learn, use zero-based indexing by default.

### 3.6. Deep Learning Techniques

To model sentiment patterns in poultry-production discourse, four predictive models were employed: CNN, DistilBERT, RoBERTa, and the proposed PoultryLeX-Net. These models were chosen for their demonstrated effectiveness for text categorization tasks. Following prior work (Salameh et al., 2025), a CNN was used in this study to extract hierarchical features from textual data for classification. CNNs are commonly employed in SA due to their ability to capture local n-gram features and hierarchical patterns in text (Patankar, 2025). CNNs easily detect sentiment-bearing phrases by applying multiple convolutional filters over word-embedding matrices. Additionally, the DistilBERT model, as implemented in (Khan et al., 2025), was employed in this study to leverage its proven effectiveness for text-based classification tasks. DistilBERT is a compressed yet semantically robust variant of BERT. It preserves the bidirectional contextual encoding feature of transformer systems while reducing model size and computational overhead (U. Sharma et al., 2025). DistilBERT's efficiency makes it suitable for large-scale sentiment classification without significantly reducing performance.

Following the approach of (Yadalam et al., 2025), the study employed the RoBERTa model, utilizing its pre-trained capabilities for robustness. RoBERTa model also improves BERT by using optimized pre-training procedures, eliminating next-sentence prediction, and training on substantially larger corpora (Hassan et al., 2025). Its capacity to collect deep bidirectional contextual information allows it to handle subtle emotion expressions and domain-specific language variances found in poultry production records. Finally, PoultryLeX-Net integrates four core components: a domain-adaptive transformer-based embedding layer (Alqarni et al., 2025), a dual-stream contextual encoder that captures both global semantics and local lexical cues (Goud and Garg, 2025), a gated-attention cross-fusion module that unifies these representations (Dubey et al., 2025), and a classification projection layer that generates multi-dimensional lexical assessments (Khan et al., 2025). **Fig. 2** illustrates the overall PoultryLeX-Net architecture along with its global attention layer.

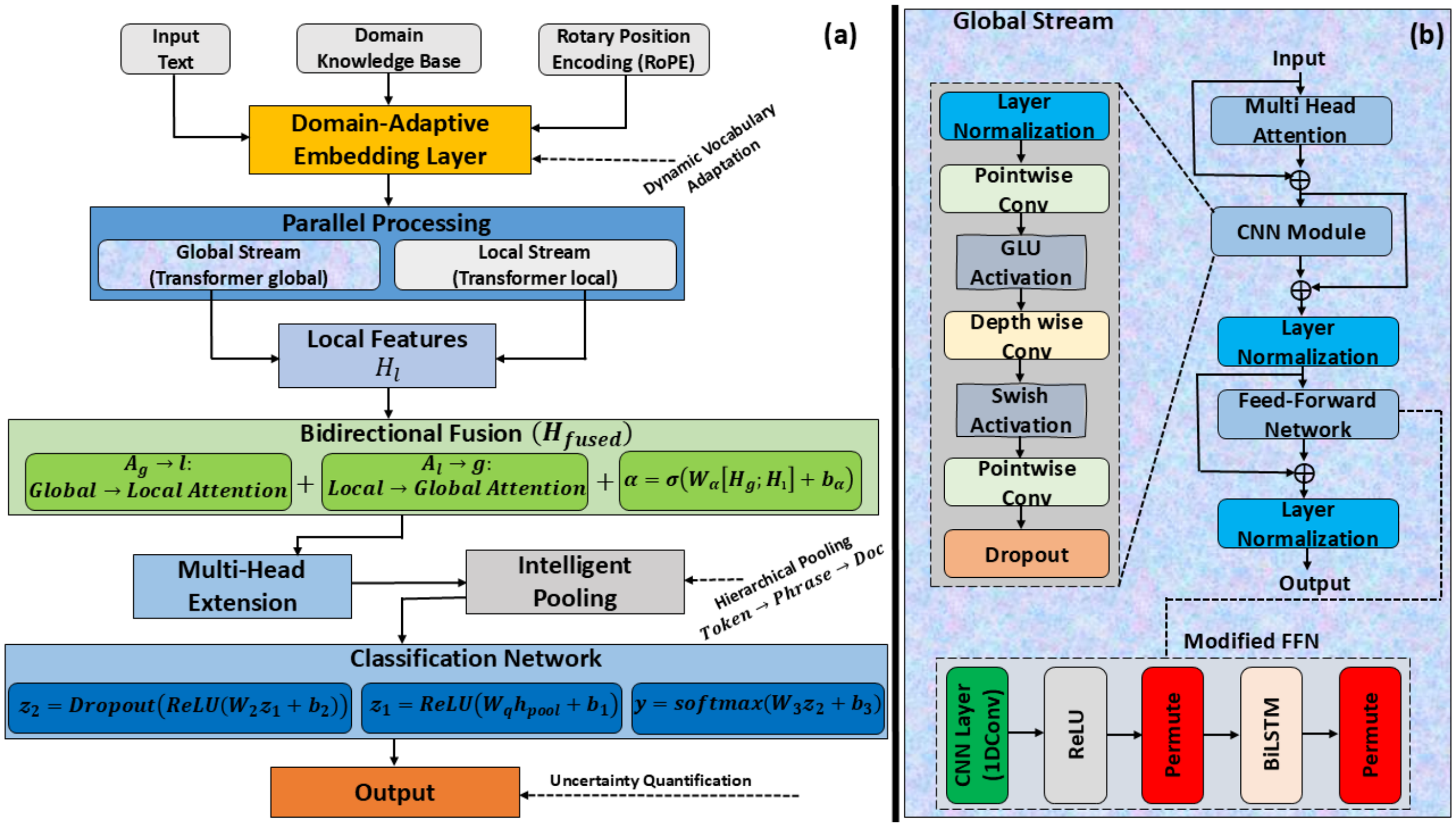

**Fig. 2.** (a) Detailed PoultryLeX-Net architecture illustrating lexicon-guided and contextual dual-stream transformer components, and (b) operation of the global contextual stream layer modeling long-range semantic dependencies within the network.

Given an input sequence of tokens $X = \{x_1, x_2, \cdots, x_n\}$, the domain-adaptive embedding layers produce equation 7.

$$E = Embed_{domain}(X) + PE_n \tag{7}$$

Where $Embed_{domain}$ is the data-adapted transformer embeddings, and $PE_n$ is the positional encodings. E is the token representation combining domain-specific embeddings with $PE_n$. The global context stream is computed using equation 8, whereas the local context stream is derived using equation 9.

$$H_g = Transformer_{global}(E) \tag{8}$$

$$H_l = Transformer_{local}(E) \tag{9}$$

Each transformer block in both streams follows the standard self-attention mechanism (Ganjalipour et al., 2025) in equation 10.

$$Attention(Q, K, V) = softmax\left(\frac{QK^T}{\sqrt{d_K}}\right)V \tag{10}$$

The gated cross-attention mechanism is defined through equations 11 to 14.

$$A_{g\rightarrow\iota} = softmax\left(\frac{H_g W_q (H_\iota W_K)^T}{\sqrt{d}}\right) H_\iota W_v \quad (11)$$

$$A_{\iota\rightarrow g} = softmax\left(\frac{H_\iota W_q' (H_g W_K')^T}{\sqrt{d}}\right) H_g W_v' \quad (12)$$

$$\alpha = \sigma\left(W_\alpha\left[H_g; H_\iota\right] + b_\alpha\right) \quad (13)$$

$$H_{fused} = \alpha \cdot A_{g\rightarrow\iota} + (1-\alpha) \cdot A_{\iota\rightarrow g} \quad (14)$$

Where $\sigma$ is the sigmoid function and $W$ are learnable parameters. Finally, the classification stage is computed using Equations 15 to 18.

$$h_{pool} = Pool\left(H_{fused}\right) \quad (15)$$

$$z_1 = ReLU\left(W_q h_{pool} + b_1\right) \quad (16)$$

$$z_2 = Dropout\left(ReLU(W_2 z_1 + b_2)\right) \quad (17)$$

$$y = softmax(W_3 z_2 + b_3) \quad (18)$$

The gated cross-attention mechanism within the fusion model is represented by equation 19.

$$GatedCrossAttention\left(H_g, H_\iota\right) = \sum_{i=1}^{n} \lambda_i \cdot CrossAtt_i\left(H_g, H_\iota\right) \quad (19)$$

Where $\lambda_i$ are learned gating parameters and $CrossAtt_i$ represents individual cross-attention heads. The modular design allows for easy adaptation to other domains, while the domain-adaptive embedding ensures relevance to specialized terminology.

### 3.7. Performance Evaluation Metrics

Precision, recall, F1-score, accuracy, and the area under the receiver operating characteristic curve (AUC-ROC) were utilized to evaluate model robustness. Confusion matrices were used to visualize classification behavior based on True Positives (TP), True Negatives (TN), False Positives (FP), and False Negatives (FN). Accuracy represents the proportion of correctly identified samples, whereas precision is the fraction of genuine positives among predicted positives (Al-henaki and Al-khalifa, 2025). The F1-score gives a balanced harmonic mean of accuracy and recall (Gao and Fang, 2025), which is especially valuable when there is a class imbalance (Sharma et al., 2025). The ROC curve compares the True Positive Rate (TPR = Recall) to the False Positive Rate (FPR) across thresholds, with AUC representing overall discrimination performance (Saranya and Saran A., 2025). For multi-class sentiment classification, a one-vs-rest (OvR) approach was employed, comparing each class against all others, enabling rigorous assessment of each class's discriminative ability, as commonly applied in classification studies. Equations 20 to 24 establish the evaluation metrics.

$$Accuracy = \frac{TP+TN}{TP+TN+FP+FN} \quad (20)$$

$$Precision = \frac{TP}{TP+FP} \quad (21)$$

$$Recall = \frac{TP}{TP+FN} \quad (22)$$

$$F1_{score} = 2 \times \frac{Precision \times Recall}{Precision + Recall} \quad (23)$$

$$FPR = \frac{FP}{FP+TN} \quad (24)$$

These metrics allow thorough, balanced, and interpretable evaluation of model performance, facilitating accurate sentiment classification across varied textual inputs.

## 4. Results

This section presents the findings from SA, LDA topic modeling, and DL classifiers, demonstrating how these approaches uncover major themes, sentiment patterns, and predictive insights in poultry production discourse.

### 4.1. Performance of Sentiment Analysis

The performance of the SA includes term-document matrix (TDM), word-cloud outputs, and sentiment score distributions. The results show how emotional tone varies across poultry-related talks and highlight the key lexical factors that shape sentiment polarity. The TDM indicated that the most frequently occurring words in the corpus focus on key aspects of poultry production. These findings highlight the relevance of feed management and nutritional optimization. Breeding procedures and performance monitoring are also emphasized. Together, they demonstrate the study's emphasis on improving poultry production efficiency and productivity. **Fig. 3** shows the top 20 keywords and their frequency.

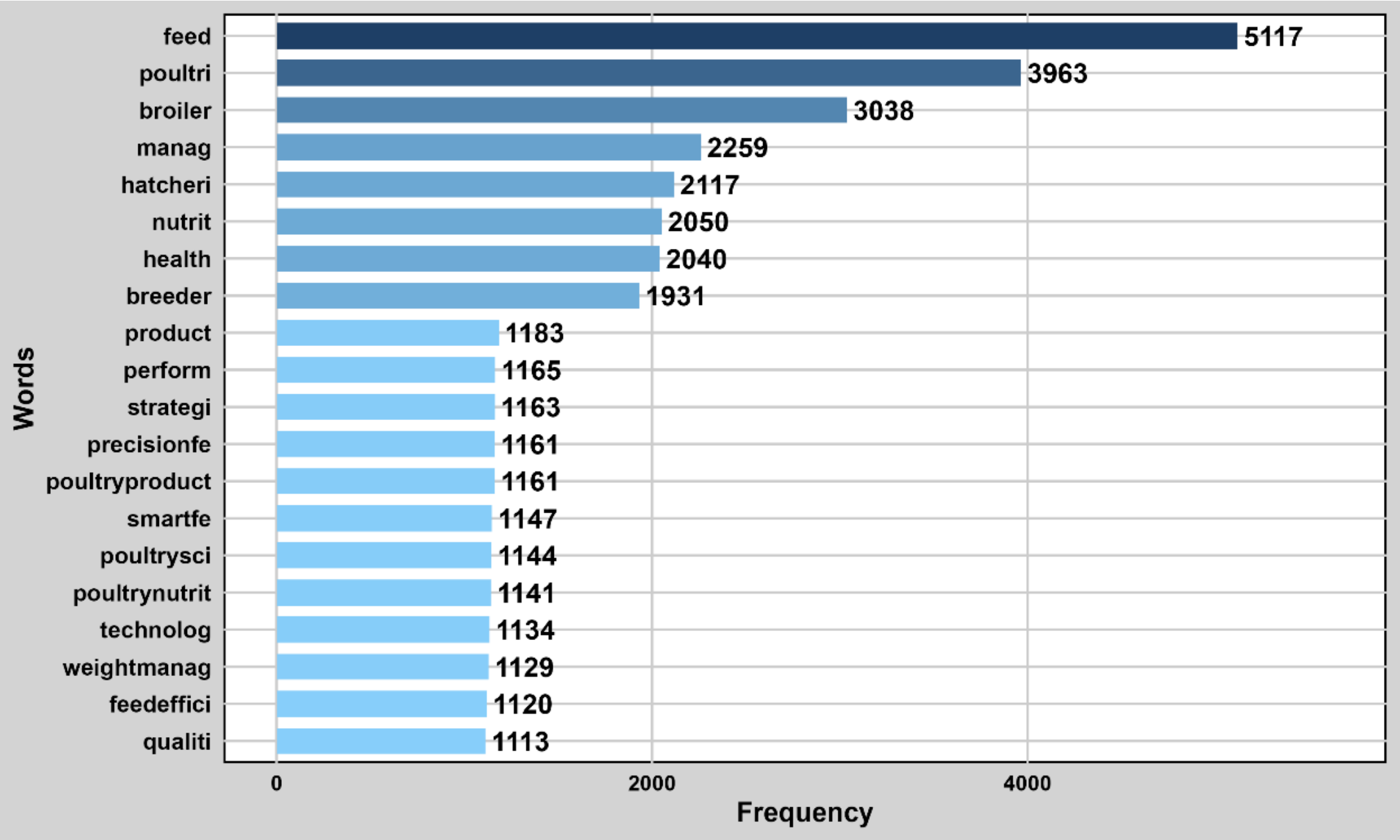

**Fig. 3.** Distribution of the top 20 most frequent terms in the poultry-related text corpus, highlighting dominant vocabulary used in sentiment and thematic analysis.

The word cloud visualization shows the most frequently used words in corpus, showing significant themes and operational goals (see **Fig. 4**). Larger font size terms like "hatchery", "poultry", "feed", "health", "feeding", "management", "smartfeeding", "poultrynutrition", "breeder", "feedconversion", "nutrition", and "weightmanagement" indicate key areas of concentration. Terms like "feed" and "health" refer to basic production requirements. Others such as "smartfeeding", "weightmanagement", and "feedconversion" are terms that refer to specific procedures that attempt to improve efficiency and performance. Their predominance across the corpus emphasizes the practical and strategic aspects of poultry management, as well as how language reflects both normal operations and targeted interventions for optimal outcomes.

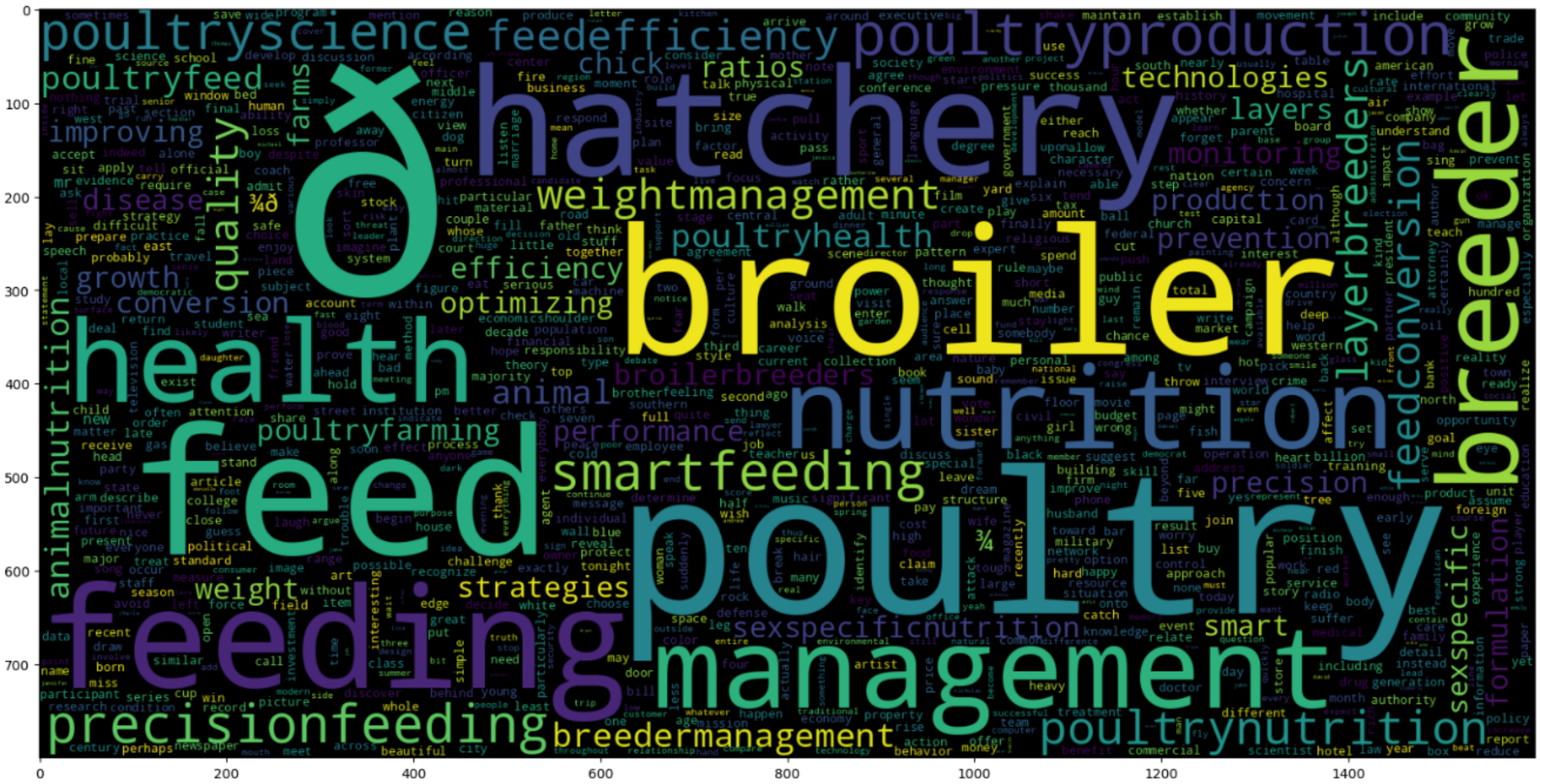

**Fig. 4.** Word cloud visualization of prominent terms in the poultry-related corpus, illustrating relative term importance based on frequency.

SA of the dataset revealed a majority of positive and trust-related terms, indicating overall positive impressions. The study found 24,144 positive and 13,099 trust-related events, indicating confidence and optimism in the discussion. Other notable feelings included anticipation (9,179), joy (5,797), fear (5,237), negative (6,159), sadness (4,084), anger (3,392), surprise (3,635), and disgust (2,863). These findings show that, while positive and trust sentiments dominate, the prevalence of negative and fearful expressions highlights the dataset's diverse emotional context. **Fig. 5** shows the distribution of sentiment scores. Word clouds show essential operational tactics, whereas sentiment scores reflect generally positive and trusting feelings, indicating confidence in production practices as well as notable negative and fear expressions, emphasizing areas that require attention.

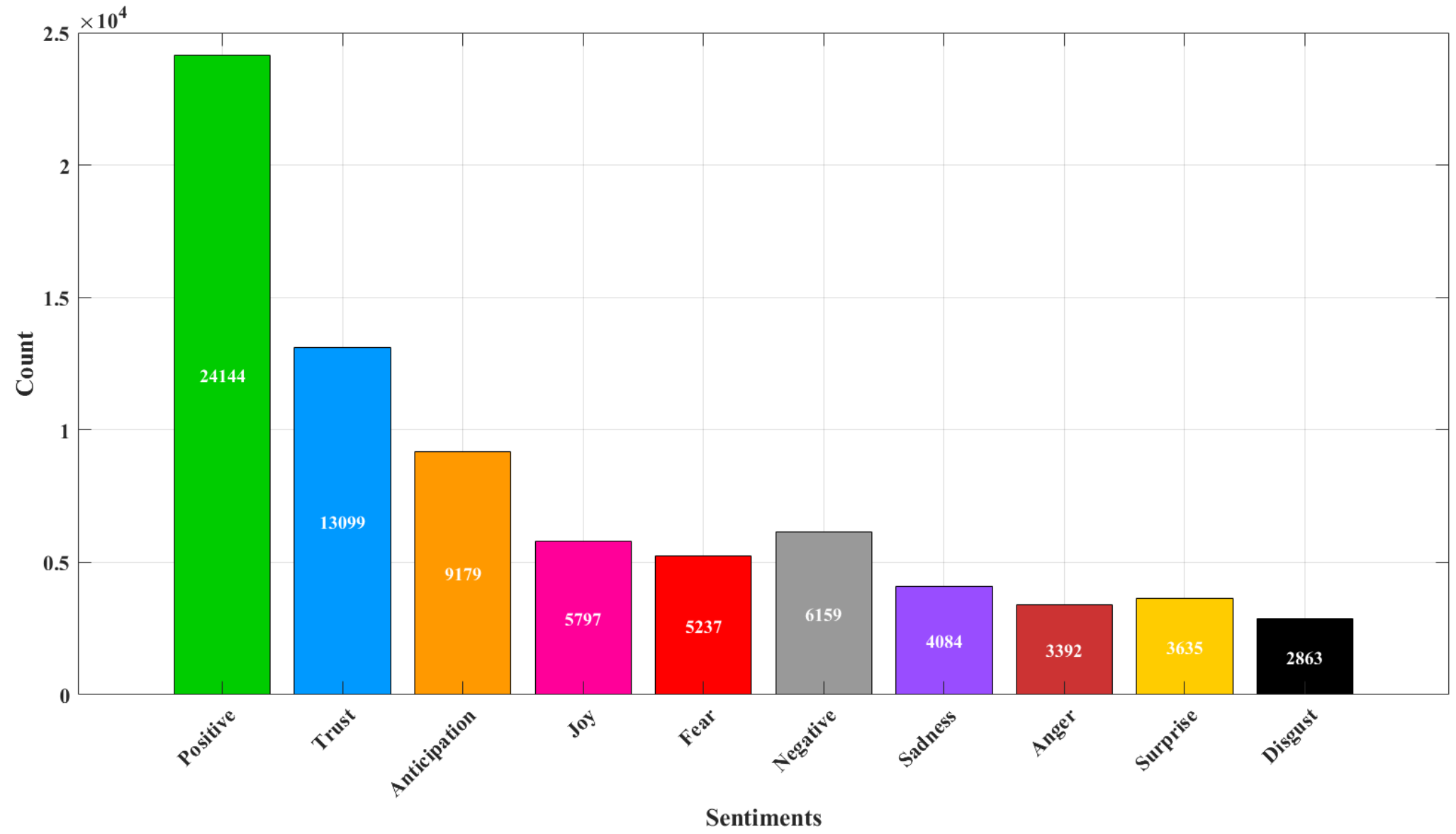

**Fig. 5.** Distribution of sentiment scores across the poultry-related text corpus, showing the prevalence of emotion classes.

### 4.2. Performance of Latent Dirichlet Allocation Modeling

The LDA topic modeling analysis identified **five distinct themes** within the poultry text corpus (Table 4), reflecting key operational and management priorities in modern poultry production. Specifically, Topics #0 – #4 captured the dominant areas of discussion: Flock Health and Growth**,** Production Performance**,** Hatchery Management**,** Nutrition Strategies**,** and Feeding Technology and Efficiency**.** Topic #0 (Flock Health and Growth) was characterized by terms such as *broiler, breeder, health, weight, growth, monitoring,* and *management*, emphasizing flock-level health surveillance and growth optimization, whereas terms like *weightmanagement* and *animalnutrition* further highlight strategies aimed at improving bird development and maintaining overall flock condition **(Fig. 6a).** Topic #1 (Production Performance) included frequent terms such as *performance, production, poultry,* and *improving*, along with productivity-oriented keywords such as *feedefficiency, feedconversion, broilerbreeders,* and *breedermanagement*, suggesting that a major focus of the corpus is improving operational outcomes through better efficiency and performance-driven management **(Fig. 6b).** Topic #2 (Hatchery Management) was dominated by words like *hatchery, management, quality,* and *chick*, and further supported by terms such as *sexspecificnutrition, poultryfeed, smartfeeding,* and *feedconversion*, indicating an emphasis on hatchery operations, chick quality assessment, and targeted nutritional strategies to enhance hatchery success **(Fig. 6c).** Topic #3 (Nutrition Strategies) highlighted terms including *nutrition, strategy, layer, feed,* and *formulation*, reflecting strong interest in diet planning and feed formulation approaches, with additional keywords such as *sexspecific* and *smartfeeding* pointing to the growing role of precision and animal-specific feeding programs to support health and production **(Fig. 6d**). Finally, Topic #4 (Feeding Technology and Efficiency) featured dominant terms such as *feeding, technology, smart, precision, efficiency, ratio,* and *conversion*, capturing the increasing integration of smart feeding systems and precision agriculture tools to improve feed conversion, enhance farm-level efficiency, and reduce resource waste (**Fig. 6e**). Together, these themes provide a structured overview of the primary research and industry concerns represented in the corpus, showing a strong alignment between management practices, nutrition optimization, and emerging feeding technologies in poultry production.

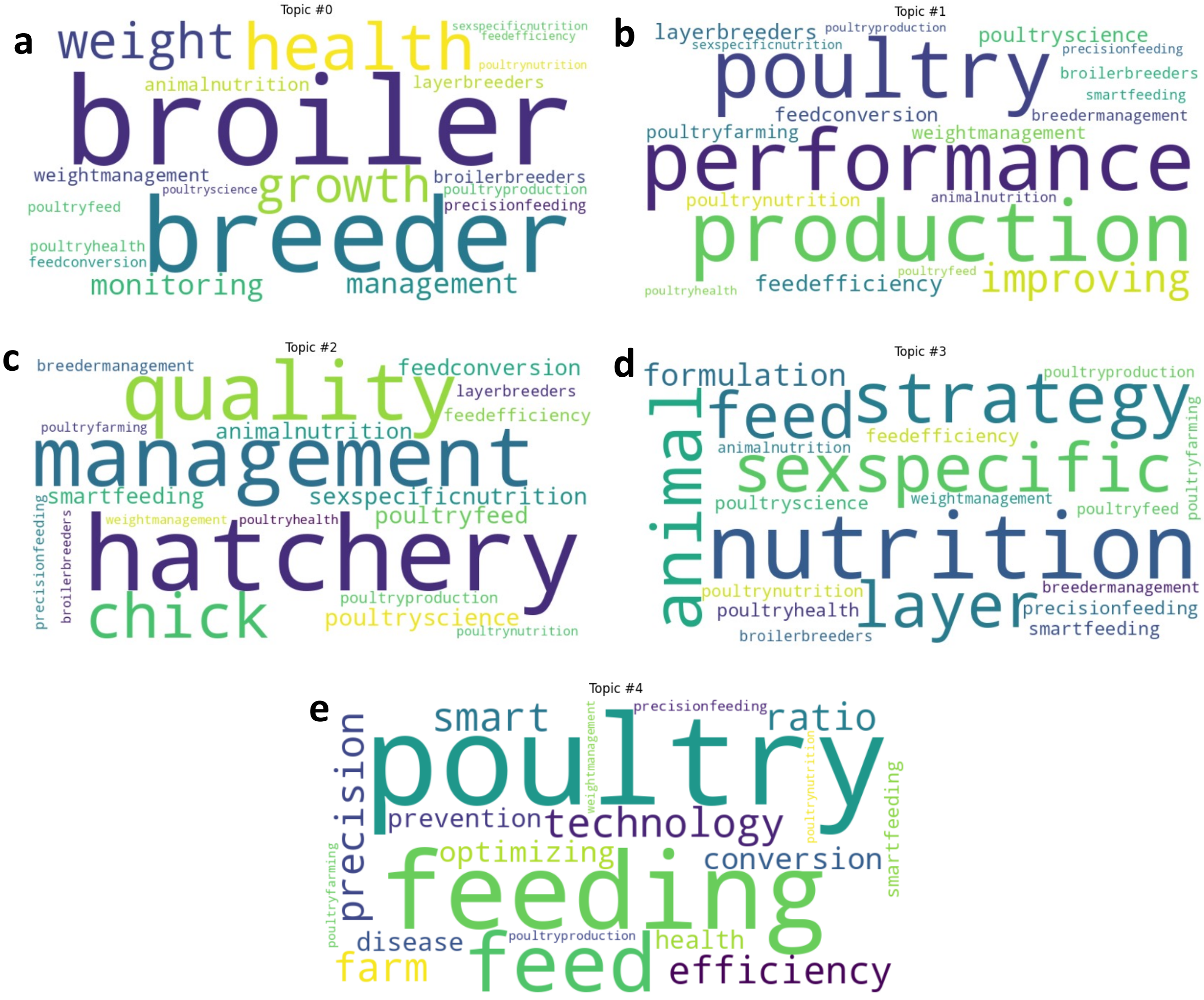

**Fig. 6.** LDA topic modeling results of the poultry-related text corpus showing five dominant themes: a) Flock Health and Growth, b) Production Performance, c) Hatchery Management, d) Nutrition Strategies, and e) Feeding Technology and Efficiency.

### 4.3. Performance of DL Models

The DL classifier was evaluated using a three-class confusion matrix for positive, neutral, and negative sentiments. A one-vs-rest (OvR) technique was used for ROC curve and AUC analysis, comparing positive and non-positive (neutral and negative) occurrences. This technique enabled rigorous assessment of the model's discriminative abilities for positive sentiment, complementing the multi-class performance metrics. **Table 5** summarizes the overall performance metrics for all four models. The CNN had the lowest scores across all metrics, DistilBERT achieved moderate improvements, while RoBERTa and PoultryLeX-Net performed at near-SOTA levels.

**Table 5.** Performance metrics of DL models for poultry sentiment classification.

| | Performance Evaluation Metrics | | | | |
|---|---|---|---|---|---|
| **Models** | **Precision** | **Recall** | **F1-score** | **Accuracy** | **AUC** |
| CNN | 0.8987 | 0.8467 | 0.8633 | 0.8932 | 0.9791 |
| DistilBERT | 0.9033 | 0.8733 | 0.8867 | 0.9134 | 0.9699 |
| RoBERTa | 0.9533 | 0.9567 | 0.9568 | 0.9583 | 0.9971 |
| PoultryLeX-Net | 0.9543 | 0.9633 | 0.9667 | 0.9735 | 0.9961 |

**Fig. 7** presents the confusion matrices for all models, offering a detailed class-wise view of prediction behavior, common errors, and the steady improvement from the CNN baseline to PoultryLeX-Net. The CNN model (**Fig. 7a**) establishes the baseline. While the overall performance is competitive, its confusion matrix shows noticeable confusion between positive and neutral classes, suggesting that the CNN struggles to capture subtle contextual dependencies and tends to misinterpret mild or indirectly expressed sentiment; an expected limitation for convolution-based text models. In comparison, DistilBERT (**Fig. 7b)** improves sentiment separation, showing fewer cross-class errors and reduced confusion between positive and neutral sentiments. This reflects its stronger contextual representation through bidirectional embedding. This improvement reflects DistilBERT's ability to learn richer contextual representations through bidirectional embeddings, allowing more accurate semantic differentiation in poultry-related text. A more substantial performance gain is observed with RoBERTa (Fig. 7c), whose predictions are more strongly concentrated along the diagonal, indicating consistently robust classification across all sentiment categories and improved handling of nuanced expressions.. Finally, PoultryLeX-Net (**Fig. 7d**) delivers the best overall results, with the most balanced confusion matrix and minimal misclassification across all classes. The near-perfect diagonal dominance suggests that PoultryLeX-Net not only learns general sentiment cues effectively but also captures domain-specific expressions more reliably, likely supported by its efficient attention-flow mechanism and domain-adaptive fine-tuning. Overall, these confusion matrices visually confirm that as contextual modeling becomes stronger and more domain-aware, misclassifications, especially between closely related sentiment classes like positive and neutral, can be decreased substantially, with PoultryLeX-Net providing the most stable and accurate sentiment classification performance for poultry-related text.

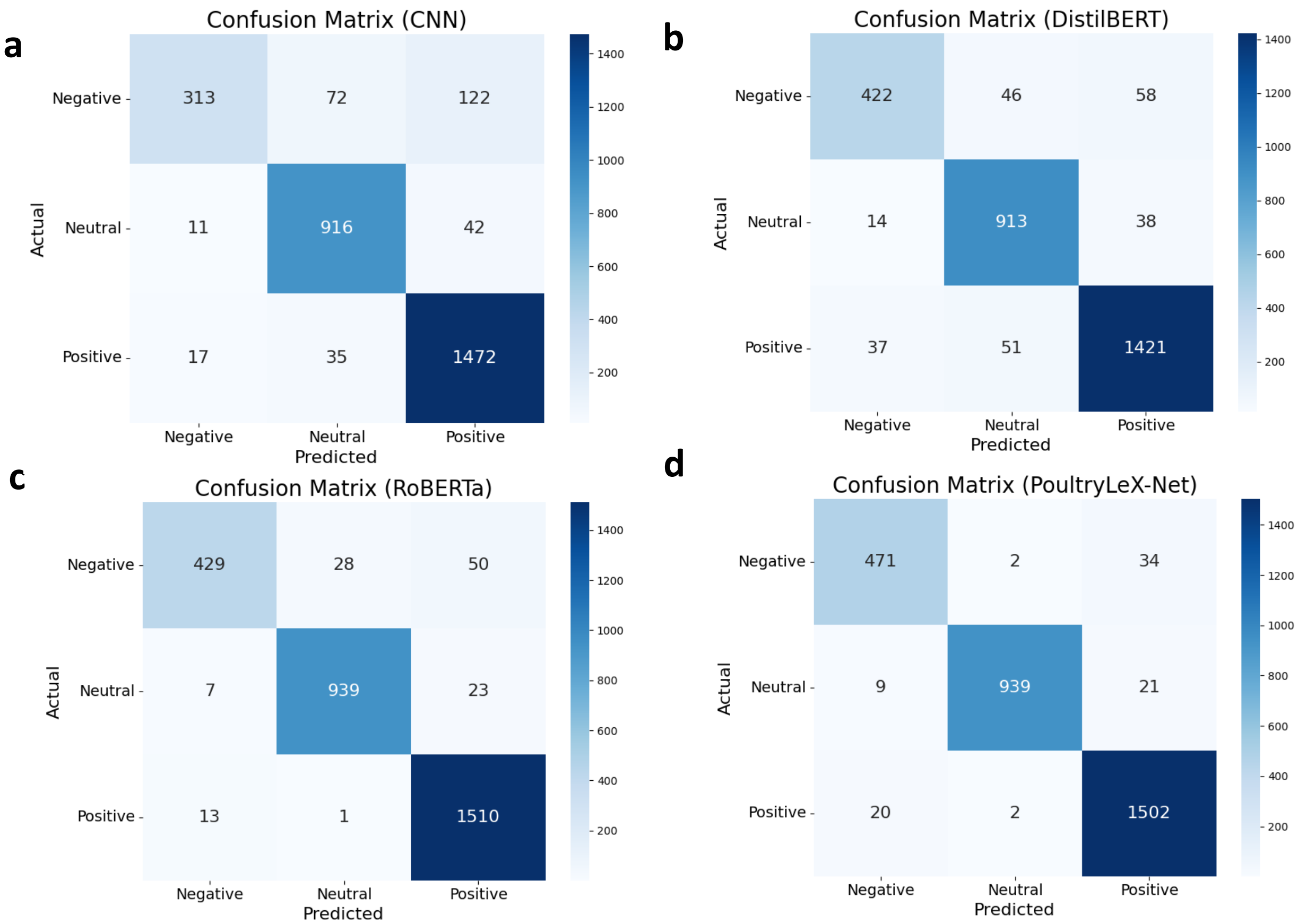

**Fig. 7.** Confusion matrix of (a) CNN, (b) DistilBERT, (c) RoBERTa, and (d) PoultryLeX-Net showing classification performance across sentiment classes.

**Fig. 8** presents a clear visual comparison of the performance under a OvR strategy, where the multi-class problem is reduced to binary (Positive vs. Non-Positive). This highlights each model's ability to distinguish Positive sentiment and confirms the performance trend observed in the multi-class results. Overall, the ROC curves show a steady improvement in discriminative power as the models move from conventional deep learning to transformer-based architectures. In **Fig. 8a**, the CNN model achieves an AUC of 0.9791, indicating that it can distinguish positive sentiment reasonably well. However, this performance is mainly driven by pattern-based feature learning and may still struggle when sentiment is expressed indirectly or through more complex poultry-specific language. In **Fig. 8b**, DistilBERT achieves an AUC of 0.9699, which remains strong and demonstrates reliable separation while maintaining a lightweight, computationally efficient design. Although its AUC is slightly lower than the CNN in the positive vs. non-positive setting, DistilBERT demonstrates comparable classification performance. Given its architectural design and computational efficiency, it remains a practical option when speed and resource efficiency are priorities (Khan et al., 2025). The strongest class separation is observed in **Fig. 8c**, where RoBERTa achieves an AUC of 0.9971, demonstrating near-perfect discrimination and confirming its ability to capture subtle semantic relationships and domain-specific sentiment cues common in poultry-related texts. Finally, **Fig. 8d** highlights the performance of PoultryLeX-Net, which achieves an AUC of 0.9961, closely matching RoBERTa in ROC behavior while also outperforming it in overall accuracy and F1 score across the full multi-class task. This suggests that PoultryLeX-Net not only separates positive sentiment extremely well but also recognizes more nuanced or mixed sentiment boundaries. Taken together, these ROC–AUC results strengthen the overall findings of this study by showing that transformer-based models, and particularly PoultryLeX-Net, provide more stable and reliable sentiment discrimination for poultry-focused corpora, making them well-suited for real-world decision-support and monitoring applications in the poultry domain.

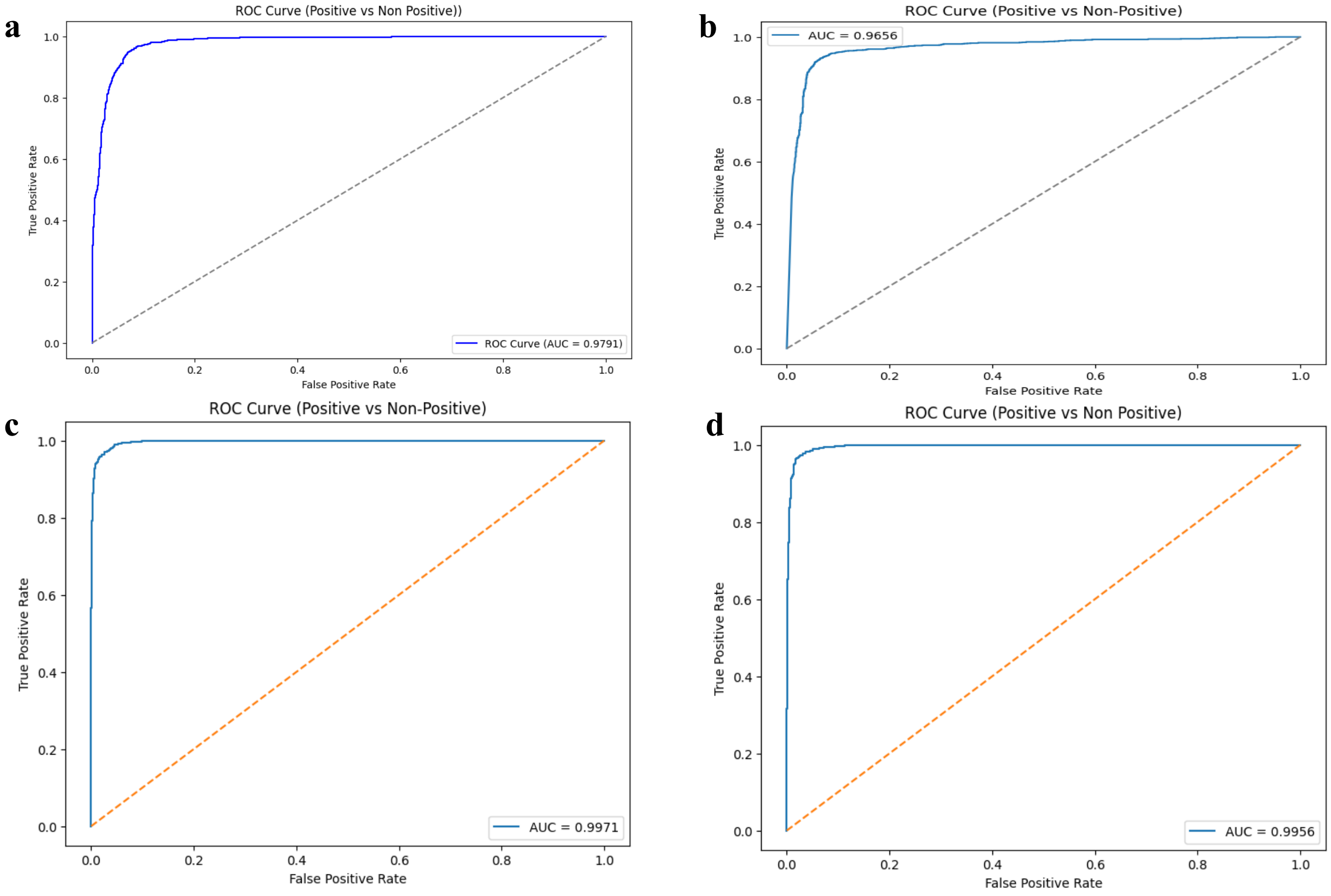

**Fig. 8.** ROC curve of (a) CNN, (b) DistilBERT, (c) RoBERTa, and (d) PoultryLeX-Net showing class discrimination performance.

Across all evaluated models, PoultryLeX-Net consistently outperformed the baselines, achieving the highest accuracy (0.9735), F1-score (0.9667), and AUC (0.9961). RoBERTa also demonstrated strong performance, with near-perfect ROC separation and high multi-class classification metrics, while DistilBERT offered robust and computationally efficient results. The CNN model provided a reliable baseline but showed more misclassifications, particularly between positive and neutral classes. These results confirm that PoultryLeX-Net delivers superior generalization, precision, and robustness for multi-class sentiment classification of poultry-related corpus.

## 5. Discussion

The integrated study presents a consistent linguistic and computational portrayal of stakeholder viewpoints in poultry production. The SA indicated that discussions are largely optimistic, with positive language suggesting that participants express trust in the information being exchanged. Albeit there are pockets of concern about issues such as disease risks, feed variability, and production uncertainty. This emotional duality emphasizes both industry confidence and ongoing issues of concern. Topic modeling helped to clarify these patterns. The five LDA-derived themes strongly aligned with modern poultry systems' main operating pillars. Their prevalence indicates the online discourse is heavily focused on efficiency-driven and technology-enabled management approaches. The prevalence of terms such as feed optimization, precision technology, and performance monitoring reveals a sector increasingly shaped by data-centric and science-based decision-making. The DL classification strengthened these findings by accurately classifying sentiment polarity across the corpus. PoultryLeX-Net achieved the highest accuracy and AUC, outperforming RoBERTa, DistilBERT, and CNN. PoultryLeX-Net's superior classification of positive, neutral, and negative sentiments demonstrates the feasibility of advanced transformer-based designs for domain-specific sentiment monitoring. These findings reveal that integrating lexical analysis, topic modeling, and DL offers a strong framework for understanding industry perceptions and leading data-informed changes in poultry production systems. **Table 6** compares the models to SOTA classification algorithms.

**Table 6.** Comparison of SOTA agricultural text analytics studies.

| Study | Dataset | Geographical Location | Methodology | Prediction Accuracy | Comparative Insights and Contributions |
|---|---|---|---|---|---|
| Domínguez et al. (2024) | X posts on agricultural policy issues | Valencia, Spain | NLP-based sentiment and topic analysis | — | Showed feasibility of using social media discourse to gauge public reaction to agricultural policy changes. However, it did not apply DL classification. |
| Bagheri et al. (2023) | Textual records from food production in a regional farm context. | Oklahoma Panhandle, United States | SA and DL classification | Classification results: Accuracy of 0.6543 | Identified production-related themes and general knowledge-discovery insights in sustainable farming. However, it did not apply DL models for sentiment classification |
| Zhu et al. (2025) | Historic livestock price data and unstructured news and social-media data | Cheongju, South Korea | SA and attention-based LSTM for price forecasting | Classification results: F1-score of 0.7941 | Demonstrated how sentiment signals from unstructured text can improve market forecasting under external shocks, however, it did not perform thematic modeling |
| Duncan et al. (2024) | Transcribed semi-structured interviews with livestock veterinarians on mental wellbeing | United Kingdom | Text mining + SA | — | Provides thematic and sentiment insights related to mental wellbeing and stressors in livestock veterinary practice. However, it did not operationalize automated domain-adaptive sentiment classification |
| Proposed study | Large-scale textual data from | | Dual-stream transformer, gated | Accuracy 0.9735, F1- | Introduces **PoultryLeX-Net**, a domain-adaptive DL |

| | | | |
|---|---|---|---|
| poultry production environments | cross-attention, domain-adapted embeddings, multi-class sentiment classification | score 0.9667, AUC 0.9961 | outperforming SOTA and supporting precision management and decision-making in poultry operations. |

## 6. Future Works

In future, the applicability of PoultryLex-Net could be expanded by using other data sources (e.g., Youtube, WhatsApp). While social media provides timely insights into public sentiment and production practices, it may introduce selection bias, as content is generated by a subset of users who may not represent the broader poultry community (Paramesha et al., 2023). Influential users can disproportionately shape discussions, and online discourse may overlook technical or clinical aspects of poultry management (AminiMotlagh et al., 2022). Additionally, although topic modeling and SA capture general patterns, they cannot fully replace expert veterinary evaluation, particularly for interpreting flock health, disease, and production efficiency. Future studies could integrate multiple social media platforms, field surveys, and expert validation to provide a more comprehensive and representative view of the poultry production sector. Incorporating systematic ground-truth validation through farm-level data collection and expert veterinary assessment would be essential to confirm and contextualize the patterns identified through computational analysis.

## 7. Conclusion

This study combines SA, LDA topic modeling, and DL classification to provide a comprehensive assessment of stakeholder views and theme priorities in the poultry industry. The SA yielded primarily positive and trust-based discourse, indicating confidence in continued technical and management advancements. It also captured concerns about health risks, feeding efficiency, and production variability. The LDA data also showed that discussions regularly revolved around the primary operational areas of flock health, performance improvement, hatchery processes, nutrition strategies, and precision feeding technologies. These thematic patterns suggest that the sector is becoming progressively shaped by data-driven management practices and technology-enabled decision-making. These findings were supported by DL classification models, with PoultryLeX-Net outperforming baseline CNN, DistilBERT, and RoBERTa models in terms of sentiment class accuracy and discrimination. Together, the findings provide a strong analytical framework for monitoring industry discourse and informing precision-driven poultry production techniques.

**Acknowledgments**

This project was supported by North Carolina State University’s College of Agriculture and Life Sciences and Department of Biological and Agricultural Engineering Faculty startup fund.

**CRediT authorship contribution statement**

Stephe Afrifa: Writing – review & editing, Writing – original draft, Visualization, Validation, Software, Methodology, Funding acquisition, Formal analysis, Data curation, Conceptualization. Biswash Khatiwada: Writing – review & editing, Writing – original acquisition. Kapalik Khanal: Writing – review & editing, Writing – original acquisition. Conceptualization. Sanjay Shah: Writing – review & editing. Lingjuan Wang Li: Writing – review & editing. Ramesh Bahadur Bist: Writing – review & editing, Writing – original draft, Supervision, Project administration, Methodology, Funding acquisition, Conceptualization.

**Declaration of competing interests**

The authors declare that no competing financial interests or personal relationships could have influenced the work presented in this paper.

**Conflict of Interest:** The authors declare no conflict of interest.

## References

Al-henaki, L., & Al-khalifa, H. (2025). Enhancing Propaganda Detection in Arabic News Context Through Multi-Task Learning. *Applied Sciences*, 1–34. https://doi.org/https://doi.org/10.3390/app15158160

Alqarni, F., Sagheer, A., Alabbad, A., & Hamdoun, H. (2025). Emotion-Aware RoBERTa enhanced with emotion-specific attention and TF-IDF gating for fine-grained emotion recognition. *Scientific Reports*, 1–19. https://doi.org/https://doi.org/10.1038/s41598-025-99515-6 1

AminiMotlagh, M., Shahhoseini, H., & Fatehi, N. (2022). A reliable sentiment analysis for classification of tweets in social networks. *Social Network Analysis and Mining*, *13*(1), 1–11. https://doi.org/10.1007/s13278-022-00998-2

Bagheri, A., Taghvaeian, S., & Delen, D. (2023). A text analytics model for agricultural knowledge discovery and sustainable food production: A case study from Oklahoma Panhandle. *Decision Analytics Journal*, *9*(October), 100350. https://doi.org/10.1016/j.dajour.2023.100350

Bhattacharya, S., & Pandey, M. (2024). Developing an agriculture ontology for extracting relationships from texts using Natural Language Processing to enhance semantic understanding. *International Journal of Information Technology (Singapore)*, *0123456789*. https://doi.org/10.1007/s41870-024-01809-x

Bhattacharyya, R. (2025). Investigating the capabilities of two-stage clustering algorithms in automatically discovering categories of questions using Bloom ' s taxonomy. *Iran Journal of Computer Science*, *8*(3), 925–938. https://doi.org/10.1007/s42044-025-00255-7

Bonthu, S., Sree, S. R., & Prasad, M. H. M. K. (2025). SPRAG : building and benchmarking a Short Programming-Related Answer Grading dataset. *International Journal of Data Science and Analytics*, *20*(3), 1871–1883. https://doi.org/10.1007/s41060-024-00576-z

Bradford, T. L., & Quagrainie, K. K. (2024). Online media sentiment analysis for US oysters. *Aquaculture, Fish and Fisheries*, *4*(4), 1–10. https://doi.org/10.1002/aff2.191

Chung, M. K., Lee, S. Y., Shin, T., Park, J. Y., Hwang, S., Kim, M. H., Lee, J., Lee, K. J., Lim, H. S., Urtnasan, E., Jung, Y., & Kim, D. K. (2025). BERT and BERTopic for screening clinical depression on open-ended text messages collected through a mobile application from older adults. *BMC Public Health*. https://doi.org/10.1186/s12889-025-23337-4

Das, A. R., Pillai, N., Nanduri, B., Rothrock, M. J., & Ramkumar, M. (2024). Exploring Pathogen Presence Prediction in Pastured Poultry Farms through Transformer-Based

Models and Attention Mechanism Explainability. *Microorganisms*. https://doi.org/https://doi.org/10.3390/microorganisms12071274

Dikshit, S., Dixit, R., Tiwari, R., & Jain, P. (2025). Performant Multilingual Modulated and Multiplexed Memory Distilled Model with Adaptive Activation Ensembles. *SN Computer Science*. https://doi.org/https://doi.org/10.1007/s42979-025-04146-3

Domínguez, A. G., Roig-Tierno, N., Chaparro-Banegas, N., & García-Álvarez-Coque, J. M. (2024). Natural language processing of social network data for the evaluation of agricultural and rural policies. *Journal of Rural Studies*, *109*(June). https://doi.org/10.1016/j.jrurstud.2024.103341

Dubey, A., Ambasta, A., Soni, J., Doshi, P., Rane, M. R., & Kanani, P. (2025). A Hybrid Semantic – Rule-Based NLP Framework Integrating DFCI and MSKCC Approaches for Clinical Trial Matching Using UMLS and FAISS. *Ingénierie Des Systèmes d ' Information*, *30*(9), 2285–2295. https://doi.org/https://doi.org/10.18280/isi.300906

Duncan, A. J., Henry, M. K., & Lamont, K. (2024). Combining sentiment analysis and text mining with content analysis of farm vet interviews on mental wellbeing in livestock practice. *PLoS ONE*, *19*(5 May), 1–22. https://doi.org/10.1371/journal.pone.0304090

Erickson, M. G., Erasmus, M. A., Karcher, D. M., & Knobloch, N. A. (2019). Poultry in the classroom : effectiveness of an online poultry-science-based education program for high school STEM instruction. *Poultry Science*. https://doi.org/10.3382/ps/pez491

Gaikwad, P. P., & Venkatesan, M. (2025). DepreLex - BERT - Att - LSTM : An Advanced Framework for Automatic Clinical Depression Detection from Marathi Text. *Arabian Journal for Science and Engineering*, *50*(23), 20249–20268. https://doi.org/10.1007/s13369-025-10551-w

Ganjalipour, E., Hossein, A., Sheikhani, R., Kordrostami, S., & Asghar, A. (2025). Relation extraction with enhanced self - attention model for SOV word order languages : Persian case study. *International Journal of Data Science and Analytics*, *20*(6), 5569–5584. https://doi.org/10.1007/s41060-025-00767-2

Gao, X., & Fang, Q. (2025). Multi-granularity sentiment analysis and learning outcome prediction for Chinese educational texts based on transformer architecture. *Discover Artificial Intelligence*, *7*. https://doi.org/https://doi.org/10.1007/s44163-025-00459-7 (2025)

Garcia, B. T., Westerfield, L., Yelemali, P., Gogate, N., & Munoz, E. A. R. (2025). Improving automated deep phenotyping through large language models using retrieval - augmented generation. *Genome Medicine*. https://doi.org/https://doi.org/10.1186/s13073-025-01521-w

Goud, A., & Garg, B. (2025). A novel framework for aspect based sentiment analysis using a hybrid BERT ( HybBERT ) model. In *Multimedia Tools and Applications* (Vol. 84, Issue 29). Springer US. https://doi.org/10.1007/s11042-023-17647-1

Haq, I., Zhang, Y., & Alam, I. (2025). POS tagging of low - resource Pashto language : annotated corpus and BERT - based model. *Language Resources and Evaluation*, *59*(3), 3243–3265. https://doi.org/10.1007/s10579-025-09834-3

Hassan, E., Talaat, A. S., & Elsabagh, M. A. (2025). Intelligent text similarity assessment using

Roberta with integrated chaotic perturbation optimization techniques. *Journal of Big Data*. https://doi.org/https://doi.org/10.1186/s40537-025-01233-3

Jiang, Y., Atwa, E. M., He, P., Zhang, J., Di, M., Pan, J., & Lin, H. (2025). Computer Vision-Based Multi-Feature Extraction and Regression for Precise Egg Weight Measurement in Laying Hen Farms. *Agriculture*, 1–25. https://doi.org/https://doi.org/10.3390/agriculture15192035

Khan, H. U., Naz, A., Alarfaj, F. K., & Almusallam, N. (2025). Identifying artificial intelligence-generated content using the DistilBERT transformer and NLP techniques. *Scientific Reports*, 1–26. https://doi.org/https://doi.org/10.1038/s41598-025-08208-7

Leigh, R., Alberte, R., & Møller, S. (2025). Illuminating the care / repair nexus in the ‘ pandemic era ’, and the potential for care beyond repair in Danish poultry production. *Agriculture and Human Values*, 1173–1190. https://doi.org/https://doi.org/10.1007/s10460-024-10668-y

Lv, X., Lin, W., Meng, J., & Mo, L. (2024). Spillover Effect of Network Public Opinion on Market Prices of Small-Scale Agricultural Products. *Mathematics*, *12*(4), 1–17. https://doi.org/10.3390/math12040539

Ma, C., Tao, J., Tan, C., Liu, W., & Li, X. (2023). Negative Media Sentiment about the Pig Epidemic and Pork Price Fluctuations: A Study on Spatial Spillover Effect and Mechanism. *Agriculture (Switzerland)*, *13*(3). https://doi.org/10.3390/agriculture13030658

Mehra, V., Sood, S., & Singh, P. (2025). Understanding social media mood during global events : a sentiment and topic modeling study of FIFA 2022 Tweets Understanding social media mood during global events : a sentiment and topic modeling study of FIFA 2022 Tweets. *Engineering Research Express*. https://doi.org/https://doi.org/10.1088/2631-8695/ae1d0d

Munaf, S., Swingler, K., Brülisauer, F., O’Hare, A., Gunn, G., & Reeves, A. (2023). Text mining of veterinary forums for epidemiological surveillance supplementation. *Social Network Analysis and Mining*, *13*(1), 1–15. https://doi.org/10.1007/s13278-023-01131-7

Myneni, M. B., & Quadhari, F. S. (2025). Hybrid Transformer-Based Resume Parsing and Job Matching Using TextRank , SBERT , and DeBERTa. *SSRG International Journal of Electronics and Communication Engineering*, *12*(9), 63–71. https://doi.org/https://doi.org/10.14445/23488549/IJECE-V12I9P105

Pack, C. I., Zeiser, T., Beecks, C., & Lutz, T. (2025). KINLI: Time Series Forecasting for Monitoring Poultry Health in Complex Pen Environments. *Animals*, *15*(21), 1–27. https://doi.org/10.3390/ani15213180

Pandey, C. S., Pandey, S., Pandey, T., Pandey, S., Pandey, H., & Mishra, P. (2025). A Comparative Study of Machine Learning , Natural Language Processing and Hybrid Models for Academic Paper Acceptance Prediction : From Reviews to Decisions. *DESIDOC Journal of Library & Information Technology*, *45*(5), 439–447. https://doi.org/10.14429/djlit.21138 © 2025, DESIDOC A

Paramesha, K., Gururaj, H. L., Nayyar, A., & Ravishankar, K. C. (2023). Sentiment analysis on cross-domain textual data using classical and deep learning approaches. *Multimedia Tools*

*and Applications*. https://doi.org/10.1007/s11042-023-14427-9

Patankar, S. (2025). A CNN-transformer framework for emotion recognition in code-mixed English – Hindi data. *Discover Artificial Intelligence*, 1–13. https://doi.org/https://doi.org/10.1007/s44163-025-00400-y

Qin, H., Teng, W., Lu, M., Chen, X., Xieermaola, Y. E., Samat, S., & Wang, T. (2025). FE-P Net : An Image-Enhanced Parallel Density Estimation Network for Meat Duck Counting. *Applied Sciences*. https://doi.org/https:// doi.org/10.3390/app15073840

Rai, S. K., & Singh, J. P. (2025). Identification and analysis of Indian farmers' behavior toward adoption of new farming technologies and e-agriculture schemes through Twitter. *International Journal on Smart Sensing and Intelligent Systems*, *18*(1), 1–24. https://doi.org/10.2478/ijssis-2025-0019

Rastogi, A., Singh, R., Khan, M. Z., Aljubayri, I., & Ghabban, F. M. (2025). Patient Emotion and Sentiment Analysis Using Deep Learning. *Journal of Innovative Image Processing*, *7*(3), 582–601. https://doi.org/https://doi.org/10.36548/jiip.2025.3.001

Salameh, K., Suboh, T., Elamayreh, R., & Alhijawi, B. (2025). Deep linguistic analysis for depression in social media using RoBERTa and CNN. *International Journal of Speech Technology*, 825–836. https://doi.org/https://doi.org/10.1007/s10772-025-10218-9

Saranya, K., & Saran A., K. (2025). A Technological Framework and Analytical Approach in Developing a real-time twitter-integrated System for Rail Transit Grievance Management. *New Generation Computing*. https://doi.org/https://doi.org/10.1007/s00354-025-00298-1

Shahid, M., Amjad, M., & Muhammad, I. (2025). Leveraging CuMeta for enhanced document classification in cursive languages with transformer stacking. *Multimedia Tools and Applications*, 37327–37352. https://doi.org/10.1007/s11042-025-20681-w

Sharma, A., Vora, D., Shaw, K., & Patil, S. (2024). Sentiment analysis-based recommendation system for agricultural products. *International Journal of Information Technology (Singapore)*, *16*(2), 761–778. https://doi.org/10.1007/s41870-023-01617-9

Sharma, U., Pandey, P., & Kumar, S. (2025). Context-based sentiment analysis using a BiGRU DistilBERT fusion model for COVID-19 tweets. *Scientific Reports*, 1–12. https://doi.org/https://doi.org/10.1038/s41598-025-22929-9

Sheng, Q., & Vukina, T. (2024). Public Communication as a Mechanism for Collusion in the Broiler Industry. In *Review of Industrial Organization* (Vol. 64, Issue 1). Springer US. https://doi.org/10.1007/s11151-023-09929-7

Smith, M. L., Jung, J., Olynk Widmar, N., Ufer, D. J., Berikou, M., & Lusk, J. (2025). Quantifying interest in and sentiment of online media about greenhouse gas emissions from cattle production in the United States. *Translational Animal Science*, *9*(October). https://doi.org/10.1093/tas/txaf105

X.com. (2025). *Home / X*. https://x.com/home

Yadalam, P. K., Thaha, M., & Natarajan, P. M. (2025). An explainable RoBERTa approach to analyzing panic and anxiety sentiment in oral health education YouTube comments.

*Scientific Reports*, 1–14. https://doi.org/https://doi.org/10.1038/s41598-025-06560-2 1

Zhu, Y., Chuluunsaikhan, T., Choi, J. H., & Nasridinov, A. (2025). Integrating structured and unstructured data for livestock price forecasting: a sustainability study from South Korea. *Frontiers in Sustainable Food Systems*, *9*(July). https://doi.org/10.3389/fsufs.2025.1613616